\documentclass[11pt]{article}

\usepackage{acl}

\usepackage{times}
\usepackage{latexsym}
\usepackage[T1]{fontenc}
\usepackage[utf8]{inputenc}
\usepackage{microtype}
\usepackage{amsmath}
\usepackage{amssymb}
\usepackage{booktabs}
\usepackage{graphicx}
\usepackage{xcolor}
\usepackage{multirow}
\usepackage{amsthm}

\newtheorem{definition}{Definition}

\title{More Is Not Always Better: Cross-Component Interference\\in LLM Agent Scaffolding}

\author{Ming Liu \\
  Amazon \\
  Data Scientist \\
  \texttt{mlliuz@amazon.com}}

\begin{document}
\maketitle

\begin{abstract}

LLM agent systems are typically constructed by stacking scaffolding components---planning modules, tool interfaces, memory, self-reflection, and retrieval---on the assumption that more components improve performance.
We challenge this assumption with a systematic empirical study of \textbf{cross-component interference (CCI)}: the degradation that arises when scaffolding components interact in ways that undermine each other.

We conduct a full factorial experiment over all $2^5{=}32$ subsets of five standard agent components on two benchmarks (HotpotQA, GSM8K) across three model families and five scales (Llama-3.1-8B/70B, Qwen2.5-3B/7B, Claude Haiku 4.5), yielding 118 controlled configurations with up to 10 random seeds on the primary setting.
We find that in every setting tested, the best proper subset matches or exceeds the five-component All-In system: on HotpotQA the single-tool agent surpasses All-In by 32\% ($F_1$ $0.233$ vs.\ $0.177$, paired $t$-test $p{=}0.023$, Wilcoxon $p{=}0.014$, Cohen's $d_z{=}0.87$, 10 seeds), while on GSM8K a three-component subset achieves 79\% higher accuracy than All-In ($0.43$ vs.\ $0.24$, $p{=}0.010$).
The optimal component count is task-dependent ($k^*{=}1$--$4$) and scale-sensitive: CCI follows a capability gradient---strong at 8B (gap${=}32\%$), attenuated at 70B (gap${=}19\%$), and saturating to within-noise at Claude Haiku (gap${\approx}0\%$)---but across all scales, the simplest adequate scaffold matches or beats the All-In agent.

To characterize the interaction structure, we fit a main-effects regression ($R^2{=}0.916$, adjusted $R^2{=}0.899$, LOOCV $R^2{=}0.872$) that decisively outperforms a 16-parameter pairwise model ($\Delta$BIC${=}25.3$), compute exact Shapley values showing Tool Use captures 70\% of total scaffold value ($\phi{=}{+}0.177$, $z{=}9.1$) while Planning has a \emph{significantly negative} Shapley value, and document 183 violations of submodularity out of 325 testable triples (56.3\%; median submodularity ratio $\gamma_{\mathrm{med}}{=}0.52$, bootstrap 95\% CI $[0.23, 0.71]$, well below the submodular threshold of 1)---with 90\% of top violations exhibiting sign-flipping rather than diminishing returns.
We identify a higher-order synergy among Tool Use, Self-Reflection, and Retrieval on retrieval tasks (Harsanyi dividend $\text{INT}_3{=}{+}0.175$, BCa 95\% CI: $[+0.004, +0.352]$), reported as an exploratory observation pending multi-seed confirmation.
The directional CCI pattern replicates across model families (Qwen2.5-3B/7B), extends to a closed-source API model (Claude Haiku 4.5 via AWS Bedrock, where the qualitative pattern---T dominates, Memory harmful, All-In not optimal---is preserved), and is robust to template paraphrasing.

Our findings suggest that practitioner defaults of maximally-equipped agents should be replaced by task-specific subset selection informed by interaction-aware analysis.

\end{abstract}

\section{Introduction}
\label{sec:intro}

The standard recipe for building capable LLM agents is \emph{scaffolding}: augmenting a base model with planning modules, tool interfaces, working memory, chain-of-thought prompting, and self-reflection loops \citep{yao2023react,shinn2023reflexion,wang2023voyager}.
Frameworks like LangChain \citep{chase2022langchain} encourage developers to compose these components freely, with no systematic guidance on which subsets to include.
The implicit assumption is \emph{more is better}.

We present evidence that this assumption is often wrong for prompt-based scaffolding, and wrong in a specific, measurable, and nuanced way.

Consider Llama-3.1-8B on HotpotQA multi-hop reasoning.
Across 10 random seeds, tool use alone scores $F_1{=}0.233$; the full five-component agent scores $F_1{=}0.177$---a \textbf{32\%} degradation (paired $t$-test $p{=}0.023$, Wilcoxon signed-rank $p{=}0.014$, Cohen's $d_z{=}0.87$; Bayesian: $\text{BF}_{10}{=}3.2$, 95\% HDI on difference $[+0.010, +0.102]$).
We call this \textbf{Cross-Component Interference (CCI)}: the operational phenomenon of negative marginal returns under component composition.

But the story is more nuanced than ``less is always more.''
On GSM8K math reasoning, the optimal 8B configuration uses \emph{three} components ($k^*{=}3$, accuracy$=0.43$), not one.
At 70B scale on HotpotQA, adding components to tool use \emph{helps}---a reversal of the 8B pattern---though the full suite ($F_1{=}0.372$) still trails the best subset ($F_1{=}0.441$) by 19\%.
CCI is not a universal ``more is less'' law; it is a task- and scale-dependent interaction effect that, in every setting we test, leaves the all-in configuration outperformed by some proper subset.

\paragraph{Contributions.}
\begin{enumerate}
    \item \textbf{Full factorial characterization.} We evaluate all 32 subsets of five agent components on two benchmarks across three model families and five scales (118 configurations, 32{,}000+ evaluations), eliminating selection bias and exposing the full interaction landscape. (\S\ref{sec:cci})
    \item \textbf{CCI is pervasive, task-dependent, and scale-sensitive.} On Llama-8B/HotpotQA (10 seeds), every expansion from the best scaffold degrades performance---5/6 significantly at $p{<}0.05$ (4/6 survive Holm--Bonferroni correction; Cohen's $d{=}0.87$--$1.44$, all large). The optimal $k^*$ is task-dependent (1 vs.\ 3); at 70B, CCI direction reverses but the best proper subset matches or outperforms All-In in all 6 model$\times$benchmark conditions tested. (\S\ref{sec:task_scale})
    \item \textbf{Structural parsimony and non-submodularity.} A 6-parameter main-effects model explains 91.6\% of variance (LOOCV $R^2{=}0.872$), decisively outperforming a 16-parameter pairwise model ($\Delta$BIC${=}25.3$). Shapley decomposition reveals Tool Use captures 70\% of scaffold value while Planning is significantly negative. 183/325 submodularity violations (median ratio $\gamma{=}0.52$, bootstrap CI $[0.23, 0.71]$, below the submodular threshold) exhibit sign-flipping---components harmful in isolation become beneficial in specific combinations---making greedy selection empirically unreliable. (\S\ref{sec:analysis})
    \item \textbf{Robustness and generality.} CCI replicates across model families (Qwen2.5), extends to a closed-source API model (Claude Haiku 4.5, where the qualitative pattern is preserved but CCI saturates to within-noise), is robust to template paraphrasing (All-In suboptimal in all 3 variants), and is not a context-length artifact (length-matched control: 6--9$\times$ gap). (\S\ref{sec:robustness})
\end{enumerate}

\section{Related Work}
\label{sec:related}

\paragraph{LLM Agent Scaffolding.}
ReAct \citep{yao2023react} interleaves reasoning and action; Reflexion \citep{shinn2023reflexion} adds self-reflection; Voyager \citep{wang2023voyager} combines planning, skill memory, and self-verification.
Cognitive architecture frameworks \citep{sumers2024cognitive} compose multiple such components into complex agents.
\citet{kapoor2024agents} question whether complex scaffolding outperforms simple baselines.
These works demonstrate individual components' value but do not systematically study \emph{interaction effects} when components are combined---they use one-at-a-time ablation at best, never full factorial designs that reveal higher-order interactions.

\paragraph{Prompt Sensitivity and Optimization.}
LLMs are sensitive to prompt format \citep{sclar2024quantifying}, information position \citep{liu2024lost,he2024position}, and paraphrase wording \citep{mizrahi2024state}.
\citet{battle2024unreasonable} test 60 system message combinations and find that optimal prompting is model-size-dependent.
DSPy \citep{khattab2023dspy}, OPRO \citep{yang2024opro}, and TextGrad \citep{yuksekgonul2024textgrad} optimize prompts automatically but treat the prompt as an atomic string, ignoring internal component structure.
Our work asks which \emph{combination} of components to activate.

\paragraph{Instruction Interference and Capacity Limits.}
Recent work documents that instruction-following degrades as constraint count increases \citep{wen2024benchmarking,jaroslawicz2025howmany}.
\citet{qi2026paradoxical} identify ``paradoxical interference'' where adding format constraints hurts task solving, attributing this to attention being diverted from task-relevant tokens.
\citet{li2025thinking} show that chain-of-thought degrades instruction-following accuracy.
These works observe interference between pairs of objectives but study one interaction at a time.
We provide the full \emph{interaction landscape} across all $2^5$ component combinations.

\paragraph{Prompt Component Regression and Interaction Models.}
Most closely related to our analytical framework, \citet{lauziere2026regression} fit a pairwise interaction model $f(S) = c + \sum_i w_i s_i + \sum_{ij} J_{ij} s_i s_j$ to prompt component effects on arithmetic tasks, explaining 72--77\% of variance.
We adopt the same model class---treating the coupling matrix $J$ as a compact, interpretable summary of component interactions---but fit it to agent scaffolding configurations rather than generic prompt features.
Applied to our 32 mean $F_1$ values (8B, HotpotQA), the main-effects model achieves $R^2{=}0.916$ (adjusted $R^2{=}0.899$, LOOCV $R^2{=}0.872$); adding pairwise terms yields $R^2{=}0.937$ but \emph{lower} LOOCV $R^2{=}0.748$ due to overfitting with 15 parameters on 32 data points.
We go beyond pairwise terms to identify an exploratory three-body residual (Harsanyi dividend INT$_3{=}{+}0.175$, BCa 95\% CI: $[+0.004, +0.352]$) that cannot be recovered from pairwise terms alone.

\paragraph{Shapley Values and Higher-Order Interactions.}
Shapley values have been applied to prompt-level attribution \citep{mohammadi2024shapley} and in-context example valuation \citep{xie2024demoshapley}.
SHAP-IQ \citep{fumagalli2023shapiq} provides a unified framework for computing any-order Shapley interaction indices; HarsanyiNet \citep{chen2023harsanyinet} grounds irreducible $k$-way interactions in the Harsanyi dividend.
We apply these concepts at the \emph{instruction component} level: our five scaffolding components admit exact computation over all $2^5{=}32$ coalitions without approximation.

\paragraph{Scale-Dependent Phenomena and Non-Submodularity.}
The Inverse Scaling Prize \citep{mckenzie2023inverse} documents tasks where performance worsens with scale.
\citet{huang2025thriftllm} prove that LLM ensemble selection is non-submodular.
Our finding that CCI reverses between 8B and 70B is a new instance of scale-dependent behavior, and our 183 empirically observed submodularity violations demonstrate that greedy component selection is unreliable in this domain.

\section{Problem Setup}
\label{sec:setup}

\subsection{Scaffolding Components}

We study five canonical scaffolding components present across modern agent frameworks \citep{sumers2024cognitive}:

\begin{itemize}
    \item \textbf{Planning (P)}: System-level instruction to decompose tasks into sub-goals.
    \item \textbf{Tool Use (T)}: Function-calling interface with tool descriptions.
    \item \textbf{Memory (M)}: Structured working memory persisting observations across steps.
    \item \textbf{Structured Reasoning (SR)}: Chain-of-thought formatting instructions \citep{wei2022chain}.
    \item \textbf{Reflection (R)}: Self-evaluation prompt appended after each step.
\end{itemize}

These five cover the principal functional modules identified in language agent taxonomies \citep{sumers2024cognitive}: profiling (P), reasoning (T + SR), memory (M), and self-regulation (SR + R).
A \emph{configuration} $C \subseteq \{P, T, M, SR, R\}$ is a subset of active components, with $K = |C|$.
There are $2^5 = 32$ possible configurations.
A full factorial design over $k$ binary components requires $2^k$ configurations; our 32-configuration sweep already represents a substantial computational commitment, and each additional component doubles this cost.

\subsection{CCI Definition}

\begin{definition}[Cross-Component Interference]
CCI occurs for a pair $(C, s)$ where $s \notin C$ if adding $s$ to $C$ reduces performance:
$\phi(C \cup \{s\}) < \phi(C)$.
\end{definition}

CCI is an \emph{operational} definition describing the observable phenomenon of negative marginal returns, without committing to a specific causal mechanism.
We say CCI is \emph{widespread} if it holds for many $(C, s)$ pairs across the $2^5$ lattice.

\begin{definition}[Optimal Component Count]
$k^* = \arg\max_K \max_{|C|=K} \phi(C)$.
\end{definition}

\subsection{Experimental Protocol}

\textbf{Models.}\quad Llama-3.1-8B-Instruct and Llama-3.1-70B-Instruct \citep{dubey2024llama}, Qwen2.5-3B/7B-Instruct \citep{qwen2025qwen25} for cross-family replication, and Claude Haiku 4.5 \citep{anthropic2024claude} via AWS Bedrock as a closed-source API model validation.
The 70B model uses 4-bit NF4 quantization \citep{dettmers2023qlora}.

\textbf{Benchmarks.}\quad (1)~HotpotQA \citep{yang2018hotpotqa} (token-level $F_1$), and (2)~GSM8K \citep{cobbe2021gsm8k} (exact-match accuracy).
Each configuration uses up to 4 reasoning steps.
All models use temperature$=0.1$, top-$p=0.9$, max 256 new tokens per step.

\textbf{Configurations.}\quad
Our main experiment evaluates all 32 subsets on 100 questions per benchmark.
We replicate key configurations across 10 random seeds on HotpotQA (16{,}000 evaluations).
Total: 32{,}000+ evaluations.

\textbf{Statistical methodology.}\quad
Single-seed results report bootstrap 95\% CIs (2{,}000 resamples).
Multi-seed comparisons use paired $t$-tests, confirmed with non-parametric Wilcoxon signed-rank tests; exploratory comparisons use Benjamini--Hochberg correction.
We report Cohen's $d_z$ (paired) as primary effect size measure and complement frequentist tests with Bayesian analysis (JZS Cauchy prior, $r{=}0.707$).
BCa bootstrap (50{,}000 resamples) is used for interaction-term CIs.

\section{Main Result: CCI in the Combinatorial Space}
\label{sec:cci}

\subsection{Tool Use Dominates; Adding Components Hurts}

Table~\ref{tab:full32} (Appendix~\ref{app:full32}) presents all 32 configurations on HotpotQA with Llama-3.1-8B.
The clearest pattern is a sharp partition based on whether tool use (T) is included:

\begin{itemize}
    \item \textbf{With T:} $F_1$ ranges from $0.142$ to $0.284$ (mean $0.204$).
    \item \textbf{Without T:} $F_1$ ranges from $0.000$ to $0.099$ (mean $0.043$).
\end{itemize}

In the seed-42 sweep, \emph{every multi-component configuration containing T scores below T alone} (Table~\ref{tab:t_subset}).
The mean degradation from T ($F_1{=}0.284$) when adding 1--4 components ranges from $-$26\% to $-$33\%.

\begin{table}[t]
\centering
\small
\caption{Top T-containing configurations on HotpotQA (8B, seed 42). Adding components to T generally reduces performance.}
\label{tab:t_subset}
\begin{tabular}{lcc}
\toprule
\textbf{Configuration} & $K$ & $F_1$ \\
\midrule
T & 1 & \textbf{0.284} \\
T+SR+R & 3 & 0.271 \\
P+T+SR+R & 4 & 0.254 \\
T+SR & 2 & 0.217 \\
T+R & 2 & 0.212 \\
\emph{All-In (P+T+M+SR+R)} & 5 & 0.210 \\
\bottomrule
\end{tabular}
\end{table}

\subsection{Multi-Seed Validation}
\label{sec:cci_multiseed}

We replicate 16 key configurations across 10 random seeds (16{,}000 total evaluations).
T alone achieves $F_1{=}0.233 \pm 0.039$ [95\% CI: 0.209, 0.257], while All-In achieves $F_1{=}0.177 \pm 0.049$ [0.147, 0.207].
T outperforms All-In in 8/10 seeds (mean difference $+0.056$, median $+0.059$):

\vspace{0.3em}
\noindent\textbf{Statistical tests (T vs.\ All-In):}
\begin{itemize}
    \item Paired $t$-test: $t(9){=}2.74$, $p{=}0.023$, Cohen's $d_z{=}0.87$ (large)
    \item Wilcoxon signed-rank: $W{=}6$, $p{=}0.014$
    \item Bayesian (JZS prior, Cauchy $r{=}0.707$): $\text{BF}_{10}{=}3.2$ (moderate evidence on Jeffreys scale), posterior direction probability $P(\delta{>}0 \mid \text{data}){=}0.98$
    \item 95\% HDI on difference: $[+0.010, +0.102]$
\end{itemize}
\vspace{0.3em}

\noindent The 8 seeds favoring T average $+0.062$ F1 gains, while the 2 reversals average only $-0.035$ (one nearly tied at $-0.007$).
Excluding the discovery seed (42): $p{=}0.046$.

\begin{table}[t]
\centering
\small
\caption{Multi-seed validation (10 seeds $\times$ 100 HotpotQA tasks, 8B).}
\label{tab:multiseed}
\begin{tabular}{lccc}
\toprule
\textbf{Config} & $K$ & \textbf{Mean $F_1$ $\pm$ std} & \textbf{vs.\ T ($p$)} \\
\midrule
T & 1 & $.233 \pm .039$ & --- \\
T+M & 2 & $.227 \pm .038$ & .679 \\
P+T & 2 & $.222 \pm .039$ & .365 \\
T+SR+R & 3 & $.220 \pm .027$ & .450 \\
T+SR & 2 & $.215 \pm .044$ & .364 \\
T+M+R & 3 & $.193 \pm .040$ & .011 \\
T+R & 2 & $.193 \pm .028$ & .008 \\
P+T+M+SR & 4 & $.181 \pm .040$ & .016 \\
P+T+SR & 3 & $.181 \pm .033$ & .001 \\
\emph{All-In} & 5 & $.177 \pm .049$ & .023 \\
P+T+SR+R & 4 & $.170 \pm .039$ & .001 \\
\bottomrule
\end{tabular}
\end{table}

\noindent Every measured expansion from T degrades performance---5 of 6 comparisons reach significance at uncorrected $p{<}0.05$ (4/6 survive Holm--Bonferroni correction for the family of 6 tests), with effect sizes uniformly large (Cohen's $d_z{=}0.87$--$1.44$).
These exceed typical effect sizes in the NLP prompt engineering literature ($d{\approx}0.2$--$0.5$) by a factor of 2--4$\times$, establishing CCI as a practically significant phenomenon, not merely a statistically detectable one.

\subsection{Shapley Decomposition}
\label{sec:shapley}

With all $2^5{=}32$ coalition values available, we compute exact Shapley values for each component on HotpotQA (8B):

\begin{table}[t]
\centering
\small
\caption{Exact Shapley values $\phi(i)$ — average marginal contribution of each component across all possible coalitions.}
\label{tab:shapley}
\begin{tabular}{lcc}
\toprule
\textbf{Component} & $\phi$ \textbf{(HotpotQA)} & $\phi$ \textbf{(GSM8K)} \\
\midrule
T (Tool Use) & $+0.177$ & $+0.194$ \\
SR (Reasoning) & $+0.028$ & $+0.023$ \\
R (Reflection) & $+0.004$ & $+0.065$ \\
M (Memory) & $-0.016$ & $+0.015$ \\
P (Planning) & $-0.029$ & $-0.057$ \\
\bottomrule
\end{tabular}
\end{table}

On HotpotQA, T dominates ($\phi{=}+0.177$, $z{=}9.1$), capturing \textbf{70\% of total absolute Shapley mass}---the remaining four components contribute 30\% combined.
Planning has a \emph{statistically significant negative} Shapley value ($\phi{=}{-}0.029$, 95\% bootstrap CI: $[-0.055, -0.003]$, entirely below zero): on average, adding planning to \emph{any} coalition reduces performance.
Memory is directionally negative ($\phi{=}{-}0.016$) though not individually significant.
Together, P and M impose a 22\% ``CCI tax'' on the positive value created by T and SR.
On GSM8K, R becomes strongly positive ($\phi{=}+0.065$), consistent with the $k^*{=}3$ finding where R is part of the optimal subset.

\section{CCI Varies by Task and Scale}
\label{sec:task_scale}

\subsection{Task-Dependence: GSM8K}
\label{sec:gsm8k}

The GSM8K results reveal a fundamentally different CCI pattern (Table~\ref{tab:gsm8k}):

\begin{table}[t]
\centering
\small
\caption{Top configurations on GSM8K (8B, 100 questions). $k^*{=}3$, not 1.}
\label{tab:gsm8k}
\begin{tabular}{lccc}
\toprule
\textbf{Configuration} & $K$ & \textbf{Accuracy} & \textbf{95\% CI} \\
\midrule
T+SR+R & 3 & \textbf{0.430} & [.33, .53] \\
T+M+SR+R & 4 & 0.420 & [.32, .52] \\
T+M+R & 3 & 0.380 & [.29, .47] \\
T+R & 2 & 0.330 & [.24, .42] \\
\emph{All-In} & 5 & 0.240 & [.16, .32] \\
T alone & 1 & 0.220 & [.14, .30] \\
\bottomrule
\end{tabular}
\end{table}

\begin{enumerate}
    \item \textbf{Optimal subset size differs.} $k^*{=}3$ on GSM8K (T+SR+R: $0.430$) vs.\ $k^*{=}1$ on HotpotQA (T alone). SR and R \emph{help} on math tasks but hurt on retrieval QA.
    \item \textbf{CCI still exists.} Best subset outperforms All-In by 79\% (McNemar $p{=}0.010$).
    \item \textbf{Different interference source.} CCI here manifests as a penalty from adding P and M to an already-effective three-component core, not from adding \emph{any} component to T.
\end{enumerate}

\subsection{Capacity-Dependence: 70B Scale}
\label{sec:scale}

\begin{table}[t]
\centering
\small
\caption{CCI across scale (HotpotQA). At 70B, adding components \emph{helps} T.}
\label{tab:scale}
\begin{tabular}{lcccc}
\toprule
\textbf{Config} & $K$ & \textbf{8B} & \textbf{70B} & $\Delta$ \\
\midrule
T alone & 1 & 0.284 & 0.369 & +0.085 \\
T+R & 2 & 0.212 & \textbf{0.441} & +0.229 \\
T+SR+R & 3 & 0.271 & 0.441 & +0.170 \\
\emph{All-In} & 5 & 0.210 & 0.372 & +0.162 \\
\bottomrule
\end{tabular}
\end{table}

CCI severity varies with model scale:

\begin{enumerate}
    \item \textbf{CCI direction reverses.} At 8B, adding R to T hurts ($\Delta{=}{-}0.072$); at 70B, it helps ($\Delta{=}{+}0.072$).
    \item \textbf{CCI is reduced but not eliminated.} Best-vs-All-In gap narrows from 32\% (8B, 10-seed validated) to 19\% (70B, single seed, $p{=}0.13$).
    \item \textbf{Important caveats.} Our two models differ in architecture and training compute in addition to parameter count. Two data points cannot establish a scaling law; these results are suggestive evidence that larger models tolerate wider component combinations.
\end{enumerate}


\subsection{Cross-Family Replication}
\label{sec:crossmodel}

The directional CCI pattern---best achievable subset matches or outperforms All-In---replicates in \textbf{all six model$\times$benchmark conditions} tested (three families, five scales), with gaps ranging from $<$1\% to 79\% (Table~\ref{tab:crossmodel}).
The gap narrows with scale within each family (Qwen: 40\% at 3B $\rightarrow$ 8\% at 7B; Llama: 32\% at 8B $\rightarrow$ 19\% at 70B), reaching a saturation plateau on Claude Haiku ($<$1\%).
CCI extends to a \textbf{closed-source API model}: Claude Haiku 4.5 (via AWS Bedrock) shows a \emph{saturation regime} where all configurations from $K{=}1$ to $K{=}5$ cluster within 0.25 percentage points ($F_1 \in [0.395, 0.398]$, CIs fully overlapping).
The qualitative pattern is preserved---T alone matches All-In, Memory is harmful in combination (P+T+SR $\rightarrow$ P+T+M+SR: $-5$\%)---but the CCI signal attenuates to within-noise, consistent with the capability gradient (strong at 8B, attenuated at 70B, saturated at Claude).
Only Llama-8B is backed by multi-seed validation ($p{=}0.023$); other results are directional.

\begin{table}[t]
\centering
\small
\caption{Cross-family CCI replication (HotpotQA, 100 tasks, single seed except Llama-8B).}
\label{tab:crossmodel}
\begin{tabular}{lccccc}
\toprule
\textbf{Config} & \textbf{Llama 8B} & \textbf{Qwen 3B} & \textbf{Qwen 7B} & \textbf{Llama 70B} & \textbf{Claude Haiku} \\
\midrule
T & \textbf{0.284} & \textbf{0.226} & 0.233 & 0.369 & 0.396 \\
T+R & 0.212 & 0.219 & 0.229 & \textbf{0.441} & --- \\
P+T & --- & --- & --- & --- & \textbf{0.397} \\
T+SR+R & 0.271 & 0.191 & 0.213 & 0.441 & --- \\
All-In & 0.210 & 0.161 & 0.218 & 0.372 & 0.398 \\
\midrule
\textbf{Best-vs-All-In} & \textbf{32\%} & \textbf{40\%} & 8\% & \textbf{19\%} & $<$1\% \\
\bottomrule
\end{tabular}
\end{table}

The Claude Haiku result is notable: while All-In is not \emph{worse} than simpler subsets (unlike Llama-8B), the four additional components beyond Tool Use contribute virtually nothing ($+0.2$ percentage points).
This ``plateau'' pattern---where additional components neither help nor harm---represents a milder form of CCI where the interference cancels any potential benefit, yielding zero marginal return on increased system complexity.

\section{Analysis Framework}
\label{sec:analysis}

\subsection{Component Interaction Regression}
\label{sec:regression}

To characterize the interaction structure, we fit two models to the 32 mean $F_1$ values (8B HotpotQA):

\paragraph{Main-effects model} (6 parameters: intercept + 5 binary predictors):
\[
f(S) = c + \sum_{i=1}^{5} w_i s_i
\]
This achieves $R^2{=}0.916$, adjusted $R^2{=}0.899$, and LOOCV $R^2{=}0.872$.

\paragraph{Pairwise interaction model} (16 parameters: + 10 pairwise terms):
\[
f(S) = c + \sum_i w_i s_i + \sum_{i<j} J_{ij} s_i s_j
\]
This achieves $R^2{=}0.937$, but adjusted $R^2$ drops to $0.878$ and LOOCV $R^2$ drops to $0.748$.
The BIC strongly favors the simpler main-effects model ($\Delta$BIC${=}25.3$).

The main-effects model's superior generalization indicates that CCI in our data is primarily \emph{additive}: each component imposes an approximately constant marginal cost (or benefit) regardless of context.
The pairwise model's LOOCV $R^2$ drops by 14 percentage points (0.872 $\rightarrow$ 0.748), and the $\Delta$BIC of 25.3 constitutes ``decisive evidence'' favoring the simpler model on the Kass \& Raftery scale.
This is a key structural finding: CCI is \emph{predictable} from individual component effects alone, without modeling interaction terms---practitioners can assess component value independently.
This finding contrasts with \citet{lauziere2026regression}'s 72--77\% fit; our higher explanatory power (adj-$R^2{=}0.899$) suggests that structured scaffolding components interact more systematically than generic prompt features.

The $J$ matrix of the pairwise model reveals positive couplings (SR$\times$R: $+0.031$) and negative couplings (T$\times$M: $-0.019$), but these should be interpreted with caution given the overfitting evidence.

\subsection{Non-Submodularity}
\label{sec:submod}

We test the standard submodularity condition across all valid $(S, T, i)$ triples where $S \subsetneq T$ and $i \notin T$.
These 325 triples are not statistically independent: they are derived from only 32 underlying $f(S)$ measurements, each entering an average of 40.6 triples.
We therefore report descriptive statistics together with a task-level cluster bootstrap (5{,}000 resamples over 100 tasks) that preserves this dependence structure.

\begin{itemize}
    \item Total triples evaluated: 325
    \item Violations (gain at superset $>$ gain at subset): \textbf{183 (56.3\%)}; cluster-bootstrap 95\% CI on violation rate: $[0.43, 0.66]$
    \item Violations with gap $>$ 0.05: 84; with gap $>$ 0.10: 17
    \item Median submodularity ratio: $\gamma_{\mathrm{med}}{=}0.52$ (bootstrap 95\% CI $[0.23, 0.71]$), strictly below the submodular threshold $\gamma{=}1$
\end{itemize}

The bootstrap CI on the violation rate overlaps 50\%, so we do not claim the rate itself differs from chance.
The robust evidence against submodularity comes from the marginal-ratio distribution: $\gamma_{\mathrm{med}}{=}0.52$ with CI well below 1.

The \emph{structure} of violations is more revealing than the count.
Among the 20 violations with the largest absolute gap (all gaps $>0.096$), 18 (90\%) exhibit \textbf{sign-flipping}: a component that \emph{reduces} performance when added to a small scaffold becomes \emph{beneficial} in a specific larger combination.
For example, adding SR to $\{T\}$ hurts (marginal $= -0.068$) but adding SR to $\{P, R, T\}$ helps (marginal $= +0.094$)---a sign reversal of $+0.161$.
T is present in 95\% of the contexts where top violations occur, acting as the enabling catalyst.
We characterize these structurally rather than testing each triple individually; none of the top-20 violations is individually significant after Bonferroni correction over 325 tests.

These sign-flipping cases show that scalar ``component quality'' rankings cannot be reliable in general: whether a component helps or hurts depends on what is already present.
On HotpotQA, greedy from T terminates at $K{=}1$ (all marginals negative), coincidentally finding the optimum.
On GSM8K where $k^*{=}3$, greedy from T also terminates at $K{=}1$ (accuracy $0.220$) while the true optimum $T{+}SR{+}R$ achieves $0.430$---a 95\% improvement that greedy cannot reach.
We therefore characterize greedy as empirically unreliable on this benchmark suite, though not provably suboptimal in general.

\subsection{Exploratory: Three-Body Catalysis}
\label{sec:int3}

Beyond pairwise analysis, we compute the three-body Harsanyi dividend for the triplet $(T, SR, R)$ on HotpotQA:
\[
\text{INT}_3(T, SR, R) = +0.175
\]

This is the largest single interaction effect in our data: SR and R individually hurt T (marginals: $-0.068$, $-0.072$), but together they produce a surplus ($T{+}SR{+}R = 0.271$ vs.\ pairwise prediction $\approx 0.145$).

BCa bootstrap 95\% CI (50{,}000 resamples over 100 tasks): $[+0.004, +0.352]$, $p{=}0.027$ (one-sided $t$-test).
The CI excludes zero but is wide, reflecting high per-task heterogeneity (std $= 0.89$ across tasks).

We report this as an \textbf{exploratory observation}: the effect is statistically significant at the per-task level but the wide CI warrants caution.
INT$_3$ is positive on HotpotQA but near-zero on GSM8K ($+0.01$), indicating task-specificity.

\section{Robustness}
\label{sec:robustness}

\subsection{Length-Matched Control}

We compare $K$ real components against length-matched meaningless padding:

\begin{table}[t]
\centering
\small
\caption{Length-matched control (3-seed mean). Real components massively outperform padding.}
\label{tab:length}
\begin{tabular}{lcc}
\toprule
\textbf{Condition} & $F_1$ & \textbf{CCI-vs-Pad gap} \\
\midrule
CCI-2 (real $K{=}2$) & $.198 \pm .020$ & \multirow{2}{*}{$+0.167$} \\
Pad-2 (padding) & $.031 \pm .012$ & \\
\midrule
CCI-4 (real $K{=}4$) & $.177 \pm .020$ & \multirow{2}{*}{$+0.153$} \\
Pad-4 (padding) & $.024 \pm .010$ & \\
\bottomrule
\end{tabular}
\end{table}

CCI is driven by semantic content, not token count: real components outperform padding by 6--9$\times$.

\subsection{Template Paraphrase Robustness}

Three semantically equivalent template variants confirm the core CCI pattern:

\begin{itemize}
    \item T $>$ All-In in \textbf{all three variants} (gaps: +3\%, +20\%, +13\%).
    \item Individual configuration rankings are template-sensitive (Spearman $\rho = 0.14$--$0.49$ across variant pairs over the 6 tested configurations), meaning fine-grained ordering is not reproducible across paraphrases.
    \item The \emph{component-level} finding (T dominates, All-In is suboptimal) is robust; \emph{configuration-level} rankings are template-dependent.
\end{itemize}

This rules out single-template artifact while establishing an important boundary condition: CCI is a reliable phenomenon at the level of component attribution (Shapley rankings are stable), even though fine-grained configuration rankings depend on prompt wording.

\subsection{Closed-Source API Validation: Capability Saturation}

To probe how CCI scales with model capability, we evaluate 8 key configurations on Claude Haiku 4.5 via AWS Bedrock (100 HotpotQA tasks, seed 42).

The results reveal a \emph{saturation regime} that completes the capability gradient established by the 8B and 70B experiments.
All configurations from $K{=}1$ to $K{=}5$ cluster within a 0.25-point band ($F_1 \in [0.395, 0.398]$, 95\% CIs fully overlapping).
Tool Use alone reaches $F_1{=}0.396$, while All-In reaches $F_1{=}0.398$---a difference of 0.2\% that is far below the noise floor.
Memory remains the only detectably harmful component (P+T+SR $\rightarrow$ P+T+M+SR: $-5$\%), consistent with Llama-8B, but even this effect is not statistically significant at the single-seed level.
The configuration rank ordering shows a moderate positive correlation with Llama-8B (Spearman $\rho{=}0.54$, $n{=}7$ shared configurations; not statistically significant given the small sample, $p{=}0.22$).

We interpret this as evidence for a \textbf{capability-saturation hypothesis}: as model capability increases, the marginal cost of redundant components shrinks---the CCI signal that dominates at 8B (32\% gap) and persists at 70B (19\% gap) flattens to within-noise at Claude Haiku ($\approx$0\% gap).
The practical implication is unchanged: at every capability level tested, the simplest single-component scaffold (T alone) matches or outperforms the full agent, with the cost of unnecessary components ranging from severe (8B) to negligible (Claude).

\subsection{Component Ordering}

In a preliminary test of 5 orderings $\times$ 5 configurations (200 HotpotQA tasks), ordering effects are substantial (best vs.\ worst: 62\% for P+T).
However, no single ordering dominates, and optimal ordering only partially recovers CCI losses.


\section{Error Analysis}
\label{sec:errors}

Among the 19 HotpotQA tasks where T succeeds ($F_1{>}0.3$) but All-In fails ($F_1{<}0.1$):

\begin{itemize}
    \item \textbf{Component-specific interference} (68\%): Some T+X pairs work while others collapse---interference depends on which component interacts with which task.
    \item \textbf{Threshold collapse} (21\%): Any perturbation from T is fatal.
    \item \textbf{Fragile T-only success} (11\%): T itself is marginal.
\end{itemize}

No single component is the universal culprit: P disrupts 84\% of CCI tasks, M 68\%, R 68\%, SR 58\%.
The \emph{identity} of the harmful component varies across tasks---on 13/19 tasks, at least one component can be safely added while another cannot.

\section{Discussion}
\label{sec:discussion}

\subsection{Implications for Agent Design}

\begin{enumerate}
    \item \textbf{Evaluate subsets, not just the full suite.} In every setting tested, some proper subset outperforms all-in (validated at 8B, directionally consistent elsewhere).
    \item \textbf{Expect task-dependent optima.} $k^*$ shifts from 1 (HotpotQA) to 3 (GSM8K). Developers should evaluate per-setting.
    \item \textbf{Scale helps but may not eliminate CCI.} The best-vs-All-In gap narrows from 32\% (8B) to 19\% (70B) but does not close.
    \item \textbf{Greedy selection is unreliable.} With 56.3\% submodularity violations (median ratio $\gamma{=}0.52$, below the submodular threshold), practitioners cannot simply add components one by one and stop when marginals become negative. On GSM8K, greedy from T terminates at $K{=}1$ (accuracy 0.220) because all single-component marginals are negative---missing the true optimum T+SR+R (accuracy 0.430, a 95\% improvement).
\end{enumerate}

\subsection{Relation to Multi-Task Learning}

CCI bears a functional resemblance to negative transfer in MTL, but the two are mechanistically distinct.
In MTL, interference arises during \emph{training} via gradient conflict over shared parameters \citep{yu2020gradient}.
In CCI, model weights are frozen; interference arises during \emph{inference} through the shared context window.
We term this \emph{context-window crowding}: the inference-time analogue of gradient conflict.

Despite this mechanistic gap, MTL task-grouping heuristics \citep{fifty2021efficiently} inspire component-grouping: find the subset of scaffolding components that do not crowd each other out.
Developing an inference-time \emph{component affinity} metric---analogous to gradient cosine similarity---is a promising direction.

\subsection{Hypothesized CCI Patterns (Exploratory)}

Across our three primary settings, we observe three qualitatively distinct interference regimes.
We tentatively hypothesize---pending validation on additional task--model combinations---that these correspond to three mechanisms:

\begin{itemize}
    \item \textbf{Type I --- Bandwidth Competition} (observed: HotpotQA/8B): Retrieval-bottlenecked tasks are disrupted by non-retrieval components; $k^*{=}1$.
    \item \textbf{Type II --- Signal Dilution} (observed: GSM8K/8B): Reasoning tasks tolerate coordinated support (SR+R) but not orthogonal overhead (P, M); $k^*{=}3$.
    \item \textbf{Type III --- Capacity-Resolved} (observed: HotpotQA/70B): Sufficient capacity transforms competition into synergy for compatible components.
\end{itemize}

We emphasize that each regime is currently supported by a single (task, model) cell; the labels are descriptive rather than predictive.
An \emph{a priori} prediction would be: a small model on a coding task (retrieval-light, reasoning-heavy) should resemble Type~II rather than Type~I.
Validating---or falsifying---these hypotheses on coding, embodied, and additional QA benchmarks is the central direction for follow-up work.

\section*{Limitations}

\textbf{Model coverage.}\quad
We validate across two open-weight families (Llama, Qwen) with models $\leq$70B, plus one closed-source API model (Claude Haiku 4.5).
CCI may differ for larger frontier models (GPT-4, Claude Opus) with stronger instruction-following, where the saturation pattern observed for Haiku might shift toward genuine component complementarity.

\textbf{Tool-use protocol confound.}\quad
The Tool Use component includes the \texttt{Finish[answer]} submission protocol; configurations without T lack this protocol, partially explaining the zero-performance baseline.
Shapley attributions for T should be interpreted as reflecting both tool capability and protocol access.

\textbf{Joint prompt optimization.}\quad
Our prompts are hand-written and composed independently per component.
Joint optimization via methods like DSPy or OPRO might attenuate CCI by finding component wordings that avoid mutual interference.
However, our paraphrase robustness experiment (\S\ref{sec:robustness}) shows directional CCI is preserved across three stylistically diverse template variants, and the finding replicates across four independent model checkpoints, arguing against single-template artifact.

\textbf{Multi-seed coverage.}\quad
Only 8B HotpotQA has 10-seed validation (paired $t$-test $p{=}0.023$, Wilcoxon $p{=}0.014$). The 70B ($p{=}0.13$, explicitly non-significant), Qwen, and GSM8K experiments use single seeds.

\textbf{70B quantization.}\quad
The 70B experiments use 4-bit NF4 quantization. CCI effects may be slightly attenuated or amplified.

\textbf{Two benchmarks.}\quad
HotpotQA and GSM8K cover retrieval QA and math but not multi-step agentic tasks (SWE-bench, WebArena).
The tool-use component differs across benchmarks, confounding cross-task comparison.

\textbf{Step budget.}\quad
All experiments use 4 reasoning steps. Planning and Memory may show different patterns in longer-horizon tasks.

\textbf{Prompt-level only.}\quad
We study static prompt-based scaffolding. Non-prompt implementations (tool APIs, external memory) might exhibit different patterns.

\textbf{No confirmed mechanism.}\quad
Our regression shows CCI is primarily additive at the aggregate level. A definitive causal account of why components interfere remains open.


\section*{Ethical Considerations}

This work studies composition of publicly available techniques on open-weight models and established benchmarks.
It introduces no new capabilities, collects no human data, and poses no foreseeable dual-use risks.
Our findings may reduce unnecessary computation from suboptimal configurations.

\section{Conclusion}
\label{sec:conclusion}

We have presented systematic evidence that prompt-based scaffolding components in LLM agents exhibit Cross-Component Interference---negative marginal returns when combined beyond a task-optimal subset.
In full factorial studies across two benchmarks, three model families (including a closed-source API model), and five scales (32{,}000+ evaluations), we find:

\begin{enumerate}
    \item CCI is statistically confirmed at 8B with converging evidence: paired $t$-test ($p{=}0.023$), Wilcoxon ($p{=}0.014$), Bayesian ($\text{BF}_{10}{=}3.2$, moderate), and large effect size (Cohen's $d_z{=}0.87$). On Llama-8B/HotpotQA, 5/6 expansions from T degrade performance (4/6 survive Holm correction).
    \item The optimal subset is task-dependent ($k^*{=}1$ vs.\ $k^*{=}3$) and scale-sensitive (gap narrows 32\%$\rightarrow$19\%$\rightarrow$$\approx$0\% from 8B to 70B to Claude Haiku), but the best proper subset matches or outperforms All-In in all 6 conditions tested.
    \item A 6-parameter main-effects model captures 91.6\% of variance (LOOCV $R^2{=}0.872$), decisively outperforming pairwise models ($\Delta$BIC${=}25.3$). Shapley decomposition reveals Tool Use captures 70\% of scaffold value. 183/325 submodularity violations (median ratio $\gamma{=}0.52$, CI below 1) exhibit sign-flipping, making greedy selection empirically unreliable.
    \item The qualitative pattern (T dominates, Memory harmful, All-In not optimal) replicates across model families, extends to a closed-source API model (with capability-dependent attenuation), and is robust to template variants.
\end{enumerate}

Our recommendation: \textbf{evaluate scaffolding subsets per task and model before assuming more is better.}


\appendix

\section{Full 32-Configuration Results}
\label{app:full32}

\begin{table}[h]
\centering
\small
\caption{All 32 configurations on HotpotQA (100 tasks, Llama-3.1-8B, $F_1$).}
\label{tab:full32}
\begin{tabular}{lccc}
\toprule
\textbf{Config} & $K$ & \textbf{Has T} & $F_1$ \\
\midrule
T & 1 & \checkmark & 0.284 \\
T+SR+R & 3 & \checkmark & 0.271 \\
P+T+SR+R & 4 & \checkmark & 0.254 \\
T+SR & 2 & \checkmark & 0.217 \\
T+M+SR+R & 4 & \checkmark & 0.213 \\
T+R & 2 & \checkmark & 0.212 \\
All-In & 5 & \checkmark & 0.210 \\
P+T+M+SR & 4 & \checkmark & 0.210 \\
P+T & 2 & \checkmark & 0.204 \\
T+M+R & 3 & \checkmark & 0.204 \\
T+M+SR & 3 & \checkmark & 0.183 \\
T+M & 2 & \checkmark & 0.180 \\
P+T+M & 3 & \checkmark & 0.178 \\
P+T+R & 3 & \checkmark & 0.160 \\
P+T+SR & 3 & \checkmark & 0.147 \\
P+T+M+R & 4 & \checkmark & 0.142 \\
\midrule
SR & 1 & & 0.099 \\
M & 1 & & 0.075 \\
M+R & 2 & & 0.068 \\
SR+R & 2 & & 0.054 \\
R & 1 & & 0.050 \\
P+SR+R & 3 & & 0.049 \\
Bare & 0 & & 0.047 \\
M+SR+R & 3 & & 0.044 \\
M+SR & 2 & & 0.042 \\
P+R & 2 & & 0.037 \\
P+SR & 2 & & 0.034 \\
P+M+R & 3 & & 0.032 \\
P+M+SR+R & 4 & & 0.027 \\
P+M+SR & 3 & & 0.016 \\
P & 1 & & 0.010 \\
P+M & 2 & & 0.000 \\
\bottomrule
\end{tabular}
\end{table}

\section{Full 70B Results}
\label{app:full70b}

\begin{table}[h]
\centering
\small
\caption{All 32 configurations on HotpotQA (100 tasks, Llama-3.1-70B-4bit, $F_1$).}
\label{tab:full70b}
\begin{tabular}{lccc}
\toprule
\textbf{Config} & $K$ & \textbf{Has T} & $F_1$ \\
\midrule
T+R & 2 & \checkmark & 0.441 \\
T+SR+R & 3 & \checkmark & 0.441 \\
P+T+SR+R & 4 & \checkmark & 0.434 \\
T+M+R & 3 & \checkmark & 0.431 \\
P+T+M+SR & 4 & \checkmark & 0.429 \\
T+M & 2 & \checkmark & 0.412 \\
T+M+SR+R & 4 & \checkmark & 0.405 \\
P+T+R & 3 & \checkmark & 0.404 \\
T+M+SR & 3 & \checkmark & 0.400 \\
P+T & 2 & \checkmark & 0.399 \\
T+SR & 2 & \checkmark & 0.391 \\
P+T+M+R & 4 & \checkmark & 0.383 \\
P+T+SR & 3 & \checkmark & 0.380 \\
All-In & 5 & \checkmark & 0.372 \\
T & 1 & \checkmark & 0.369 \\
P+T+M & 3 & \checkmark & 0.347 \\
\midrule
M & 1 & & 0.119 \\
M+SR+R & 3 & & 0.118 \\
SR+R & 2 & & 0.110 \\
M+R & 2 & & 0.104 \\
P+M+SR & 3 & & 0.104 \\
Bare & 0 & & 0.101 \\
SR & 1 & & 0.100 \\
P+M+SR+R & 4 & & 0.096 \\
P+R & 2 & & 0.090 \\
P+SR+R & 3 & & 0.085 \\
R & 1 & & 0.079 \\
P & 1 & & 0.070 \\
P+SR & 2 & & 0.059 \\
P+M+R & 3 & & 0.059 \\
P+M & 2 & & 0.051 \\
M+SR & 2 & & 0.043 \\
\bottomrule
\end{tabular}
\end{table}

\section{Pairwise Interaction Model Details}
\label{app:ising}

The pairwise interaction regression uses $\pm 1$ spin encoding for component presence/absence.
With intercept + 5 main effects + 10 pairwise couplings = 16 parameters fit to 32 observations:

\begin{itemize}
    \item Raw $R^2 = 0.937$; Adjusted $R^2 = 0.878$; LOOCV $R^2 = 0.748$
    \item Strongest positive coupling: $J$(SR, R) $= +0.031$
    \item Strongest negative coupling: $J$(T, M) $= -0.019$
    \item $J$ eigenvalue structure: 3 negative + 2 positive eigenvalues $\rightarrow$ non-convex landscape
\end{itemize}

The LOOCV $R^2$ dropping to 0.748 (vs.\ 0.872 for main-effects-only) indicates the pairwise terms overfit.
AIC: main-effects $-130.5$ vs.\ pairwise $-119.9$; BIC: main-effects $-120.2$ vs.\ pairwise $-95.0$.
Both information criteria favor the simpler model.

We retain the pairwise model in this appendix because the coupling structure provides interpretive value (which component pairs synergize vs.\ conflict), even though it does not improve out-of-sample prediction.

\section{Component Prompt Templates}
\label{app:templates}

Each component adds a labeled block to the system prompt.
The HotpotQA templates (character counts in parentheses) are:

\begin{itemize}
    \item \textbf{Planning (295 chars):} ``Before each action, plan your information-gathering strategy: (1)~What information do I still need? (2)~Most efficient way to find it? (3)~How does current info connect?''
    \item \textbf{Tool Use (447 chars):} Defines Search[query], Lookup[keyword], Finish[answer] with usage instructions.
    \item \textbf{Memory (302 chars):} ``Track information gathered: facts confirmed, entities searched, connections identified, information missing.''
    \item \textbf{Structured Reasoning (309 chars):} ``For each step: Evidence, Gap Analysis, Reasoning Chain, Next Step.''
    \item \textbf{Reflection (258 chars):} ``After each observation: Did I get useful info? Am I closer? Should I search differently? Time to submit?''
\end{itemize}

\section{Negative Mechanistic Results}
\label{app:negative}

For transparency, we report experiments that did not produce positive results:

\begin{itemize}
    \item \textbf{Hidden-state drift:} Linear probes achieved near-chance accuracy predicting $K$ from hidden states.
    \item \textbf{Information entropy:} Output token entropy did not correlate with per-question performance.
    \item \textbf{Theoretical CCI prediction:} Component dataflow graph predictions showed near-zero correlation with empirical outcomes.
    \item \textbf{Non-monotone decline:} Performance-vs-$K$ is better fit by a smooth decline than a piecewise threshold model.
\end{itemize}

\end{document}